\documentclass[sigconf]{acmart}

\usepackage{amsmath}
\usepackage{algorithmic}
\usepackage[linesnumbered,ruled]{algorithm2e}
\usepackage{graphicx}
\usepackage{textcomp}
\usepackage{xcolor}
\usepackage{array}
\usepackage{booktabs}
\usepackage{threeparttable}
\usepackage{comment}

\newcolumntype{P}[1]{>{\centering\arraybackslash}p{#1}}

\usepackage[caption=false,font=normalsize,labelfont=sf,textfont=sf]{subfig}

\setcopyright{rightsretained}

\acmDOI{10.1145/nnnnnnn.nnnnnnn}

\acmISBN{978-x-xxxx-xxxx-x/YY/MM}

\acmConference[GECCO '21]{the Genetic and Evolutionary Computation Conference 2021}{July 10--14, 2021}{Lille, France}
\acmYear{2021}
\copyrightyear{2021}

\acmPrice{15.00}

\begin{document}
\title{An Efficient Evolutionary Deep Learning Framework Based on Multi-source Transfer Learning to Evolve Deep Convolutional Neural Networks}
\renewcommand{\shorttitle}{An Efficient EDL Framework Based on Multi-source Transfer Learning to Evolve Deep CNNs}


\author{Bin Wang}
\affiliation{%
  \institution{School of Engineering and Computer Science\\
  Victoria University of Wellington}
  \streetaddress{P.O. Box 1212}
  \city{Wellington}
  \postcode{6140}
}
\email{bin.wang@ecs.vuw.ac.nz}

\author{Bing Xue}
\affiliation{%
  \institution{School of Engineering and Computer Science\\
  Victoria University of Wellington}
  \streetaddress{P.O. Box 1212}
  \city{Wellington}
  \postcode{6140}
}
\email{bing.xue@ecs.vuw.ac.nz}

\author{Mengjie Zhang}
\affiliation{%
  \institution{School of Engineering and Computer Science\\
  Victoria University of Wellington}
  \streetaddress{P.O. Box 1212}
  \city{Wellington}
  \postcode{6140}
}
\email{mengjie.zhang@ecs.vuw.ac.nz}

\renewcommand{\shortauthors}{B. Wang et al.}

\begin{abstract}
Convolutional neural networks (CNNs) have constantly achieved better performance over years by introducing more complex topology, and enlarging the capacity towards deeper and wider CNNs. This makes the manual design of CNNs extremely difficult, so the automated design of CNNs has come into the research spotlight, which has obtained CNNs that outperform manually-designed CNNs. However, the computational cost is still the bottleneck of automatically designing CNNs. In this paper, inspired by transfer learning, a new evolutionary computation based framework is proposed to efficiently evolve CNNs without compromising the classification accuracy. The proposed framework leverages multi-source domains, which are smaller datasets than the target domain datasets, to evolve a generalised CNN block only once. And then, a new stacking method is proposed to both widen and deepen the evolved block, and a grid search method is proposed to find optimal stacking solutions.  The experimental results show the proposed method acquires good CNNs faster than 15 peer competitors within less than 40 GPU-hours. Regarding the classification accuracy, the proposed method gains its strong competitiveness against the peer competitors, which achieves the best error rates of 3.46\%, 18.36\% and 1.76\% for the CIFAR-10, CIFAR-100 and SVHN datasets, respectively. 
\end{abstract}

%
%



\maketitle

\section{Introduction}

\begin{figure*}[ht]
	\centering
	\includegraphics[width=0.9\textwidth]{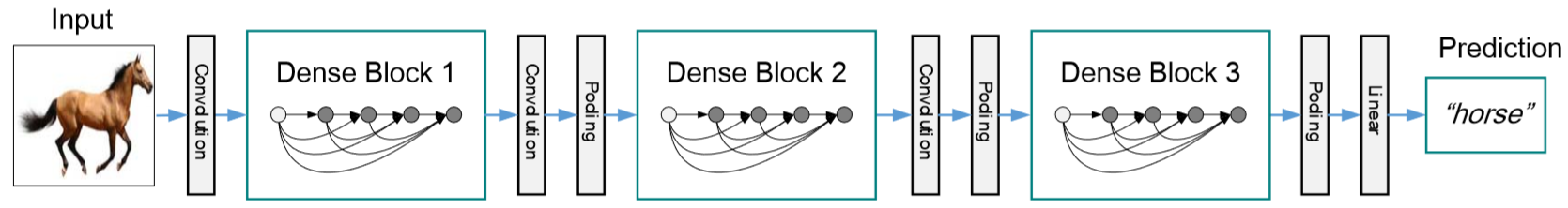}
	\caption{DenseNet architecture (Image taken from \cite{huang2017densely})}
	\label{fig:densenet_architecture}
\end{figure*}

Convolutional Neural Networks (CNNs) have been heavily investigated to solve image classification problems. Researchers have achieved the state-of-the-art classification accuracies by leveraging CNNs. In the past 10 years, to improve the classification accuracy,  CNN architectures have become more and more complex due to the depth increase and the more complex topology. For example, AlexNet \cite{krizhevsky2012imagenet} and VGGNet \cite{simonyan2014very} have tens of layers as the maximum depth and only use feed-forward connections between layers, while ResNet \cite{he2016deep} and DenseNet \cite{huang2017densely} have reached hundreds of layers and the shortcut connections have been introduced in addition to feed-forward connections. Other than that, Wide residual networks \cite{zagoruyko2016wide} and PyramidNets \cite{han2017deep} have explored different widths of the convolutional layers based on ResNet. It can be observed that CNN architectures have become extremely complex, which makes the manual design process of CNNs more difficult. The deeper and wider CNNs also need more time to train, so the trials during the manual design process take more time as well.

To avoid the tedious manual design process of CNNs, many machine learning methods have been researched to automatically design/learn CNNs. One of the successful approaches is to leverage reinforcement learning (RL) methods to search optimal CNN architectures. Outstanding works in the area are NAS \cite{zoph2016neural}, SMASH \cite{brock2017smash} and NASNet \cite{zoph2018learning}, which have achieved better classification accuracies than those of the state-of-the-art handcrafted CNNs. Another approach is to use EDL  to automatically design/learn CNNs. Promising performance has been achieved, such as GeneticCNN \cite{xie2017genetic}, DECNN \cite{wang2018hybrid}, AmobaNet \cite{real2018regularized} and MOCNN \cite{wang2020particle}. One of the most severe drawbacks is the very high computational cost, which hinders the wide usage of the above methods. For instance, NAS\cite{zoph2016neural} and NASNet\cite{zoph2018learning} found CNNs that outperformed the state-of-the-art handcrafted CNNs by having the computational cost of 22,400 GPU-days and 2,000 GPU-days, respectively. This means that NAS and NASNet are too computationally expensive to be used in small-to-medium labs and companies, which is the major hurdle for automatic methods of neural architecture search to be widely adopted. \textit{Transfer learning} and \textit{surrogate modelling} could be utilised to reduce the computational cost. In transfer learning, the knowledge, learned from solving problem in the source domain,  is applied to assist learning in the target domain. Transfer learning can reduce the computational cost in learning on the target problem by utilising knowledge learnt from the source domain. A surrogate model is a computationally cheaper method to avoid expensive evaluations in neural architecture search, which could accelerate the learning process \cite{wang2020surrogate}\cite{lu2020nsganetv2}. In this paper, an efficient EDL framework, based on transfer learning and a surrogate model called EffPNet \cite{wang2020surrogate}, will be proposed to mitigate the obstacle in order to make the automated process of designing CNNs more accessible for real-world applications.

\textbf{Goals:} The overall goal of this paper is to propose a new efficient EDL framework named EEDL. The new framework is inspired by the key idea of transfer learning. By selecting smaller datasets, in terms of both the number of examples and the image resolution, as the source domain, it is easier to learn the knowledge from the source domain than directly from the target domain. Since Particle Swarm Optimisation (PSO) is computationally inexpensive and effective for optimising a wide range of functions \cite{kennedy1995particle} \cite{eberhart1995new} \cite{shi1998modified},  the proposed EEDL framework will use PSO as the EC algorithm. However, other EC algorithms may be used as the EC algorithm in the EEDL framework. The proposed method will be evaluated on three benchmark datasets --- CIFAR-10, CIFAR-100 and SVHN. The goal is accomplished by the following contributions: 

\begin{itemize}
	\item The new EDDL framework is proposed. The source domain learning and the target domain learning are proposed to split the EDDL framework into two parts. In order to improve the generalisation of the evolve CNNs, multiple source domain datasets are proposed in the source domain learning, so the method could leverage the knowledge learned from multiple source domain datasets instead of only one source domain dataset. 
	\item A new EC-based source domain learning is proposed as a part of the overall framework. The datasets from the source domain are smaller than those of the target domain, so the evolutionary process based on the source domain could cost relatively low computational resource. 
	\item A new method of stacking the evolved CNN block to obtain CNNs with a larger capacity is proposed. Inspired by Wide residual networks \cite{zagoruyko2016wide} and NASNet \cite{zoph2018learning}, the new method of stacking the CNN block towards two directions --- depth and width, is proposed. In other words, the CNN block can be stacked to make deeper CNNs and wider CNNs to increase the model capacity. 
	\item A new target domain learning based on grid search is proposed. Based on the new stacking method, the grid search based algorithm is proposed to perform the target domain learning to find optimal CNNs in the target domain by stacking the optimal CNN block learned from the source domain. 
\end{itemize}

\section{Background and Related Work}

\subsection{DenseNet}

The major components in DenseNet \cite{huang2017densely} are dense blocks that are connected by a few convolutional layers and pooling layers with a linear layer at the end as shown in Fig. \ref{fig:densenet_architecture}. The first convolutional layer constructs the feature maps from the original images, which then are passed to the first dense block as the input feature maps. The group of a convolutional layer and a pooling layer is also known as the transition layer, which sits between two dense blocks to connect them. The convolutional layer and the pooling layer in the transition layer are fixed to a 1$\times$1 convolutional operation with a stride of 1 and a 2$\times$2 average pooling operation, respectively. For the dense block, the output feature maps are calculated according to Formula (\ref{eq:DenseNet_ouput}), where $x_{l}$ represents the output feature maps of the $l_{th}$ layer, and $[x_{0}, x_{1}, ..., x_{l-1}]$ indicates to the concatenation of the feature maps of all layers prior to the $l_{th}$ layer. $H_{l}$ is a composite function to extract the output feature maps $x_{l}$ from the concatenated feature maps. There are two hyper-parameters in a dense block --- the number of layers in the dense block and the growth rate. The \textit{growth rate} is the number of output feature maps for each layer in the dense block. DenseNet uses a fixed growth rate for all of the layers in a dense block, but it is possible to have different growth rates for different layers. 

\begin{equation}\label{eq:DenseNet_ouput}
x_{l} = F_{l}([x_{0}, x_{1}, ..., x_{l-1}])
\end{equation}

\subsection{Particle Swarm Optimisation}

Particle Swarm Optimisation (PSO) \cite{eberhart1995new} is a computational method to solve optimisation problems by iteratively moving the particles and searching for better solutions based on a given fitness evaluation function. Firstly, PSO randomly initialises a population of particles. Secondly, particles are evaluated by a fitness evaluation function, which is the problem-dependent. Thirdly, the particles in the population move to new positions according to Equations (\ref{eq:UpdateV}) and (\ref{eq:UpdateX}). In Equation (\ref{eq:UpdateV}), the velocity of the particle is updated, where $v_{id}$ is the $d_{th}$ dimension of the $i_{th}$ particle's velocity, $x_{id}$ represents the $d_{th}$ dimension of the $i_{th}$ particle's position, $P_{id}$ is the $d_{th}$ dimension of the $i_{th}$ particle's best-known position, $P_{gd}$ is the $d_{th}$ dimension of the best-known solution of the whole population (i.e. global best), $r_{1}, r_{2}$ are random numbers between 0 and 1, and $w, c_{1}$ and $c_{2}$ are the inertia weight, exploitation acceleration coefficients and exploration acceleration coefficient, respectively. In Equation (\ref{eq:UpdateX}), the new position $x_{id}(t+1)$ is obtained by adding up the velocity calculated from Equation (\ref{eq:UpdateV}) and the current position $x_{id}(t)$. In both of the equations, $(t)$ means the $t_{th}$ iteration, while $(t+1)$ indicates the $(t+1)_{th}$. Fourthly, the particles with the new positions are evaluated and the best-known position of the particle and the global best are updated. PSO iterates the above steps until a stopping criterion is met and the global best solution is returned as the output. 

\begin{equation}\label{eq:UpdateV}
v_{id}(t+1) = w * v_{id}(t) + c_{1} * r_{1} * (P_{id} - x_{id}(t)) + \\
c_{2} * r_{2} * (P_{gd} - x_{id}(t))
\end{equation}

\begin{equation}\label{eq:UpdateX}
x_{id}(t+1) = x_{id}(t) + v_{id}(t+1)
\end{equation}

\subsection{Surrogate Model of EffPNet}

The surrogate model of EffPNet \cite{wang2020surrogate} is adopted to accelerate the proposed framework in this paper because it is reliable and efficient. The surrogate model showed its reliability by achieving a prediction accuracy of more than 90\% across all generations in the experiments of EffPNet. It also demonstrated efficiency by preventing unnecessary evaluations of more than 80\% of individuals. The general idea of the surrogate model is to transform the performance prediction of a CNN to a binary classification task and then train a Support Vector Machine (SVM) \cite{cortes1995support} \cite{chang2011libsvm} to predict the comparison result of a pair of CNNs. 

The first step is to collect the data and construct a dataset of a binary classification task. The surrogate model does not need any pre-trained and pre-collected data before the evolutionary process. The data are collected from training the CNNs represented by the population at the first generation, where the training losses of all epochs, the classification accuracies of all epochs, and the best classification accuracy are recorded. From the second generation, each distinct pair in the collected data are constructed as one example of the data to fit the SVM model, where the features are comprised of the training losses and the classification accuracies of the first 10 epochs of the pair, and the class label is the comparison result of the best classification accuracies of the pair.

The constructed dataset is then used to fit an SVM model, which is used as the surrogate model to perform prediction during the evolution. Before evaluating the particle's new position, the CNN represented by the new position is trained for only 10 epochs, the training data are recorded similar to the above process of collecting data. The particle's best-known position and the training data of the current position become the pair, similar to the above process of constructing data. The surrogate model performs the prediction of whether the new position may outperform the best-known position. Only if the new position is predicted to be better, the new position will be evaluated. Otherwise, the fitness of the particle's new position is set to 0 directly. 

\section{The Proposed EEDL Framework}

\subsection{Overall Framework}

\begin{figure}[ht]
	\centering
	\includegraphics[width=.8\linewidth]{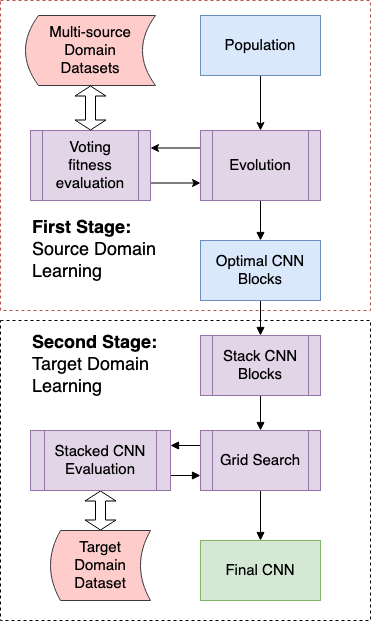}
	\caption{The proposed efficient EDL framework.}
	\label{fig:generalnet-framework}
\end{figure}

The proposed overall efficient EDL (EEDL) framework is composed of two parts/stages as shown in Fig. \ref{fig:generalnet-framework}, which are the source domain learning and the target domain learning based on transfer learning. A CNN block is evolved from the first stage of the source domain learning, and a more complex CNN is learned by stacking the CNN block from the second stage of the target domain learning as the final CNN architecture. Another benefit of introducing the two-stage learning with multi-source domain datasets is that a generalised CNN block is expected to be evolved only once. For any target domain dataset, the evolved CNN block can be used for the second stage of the target domain learning, so the first stage does not have to be repeated, which significantly saves the computational cost. 

In the source domain learning, the population is evolved and evaluated in the evolutionary process to produce optimal CNN blocks. In the proposed method, PSO is used as an example of the EC algorithm. The source domain dataset needs to be selected first, which ideally would be a much smaller dataset in terms of both the sample size and the image resolution, and would also be very similar to the target domain dataset in terms of both the image shapes and the class distribution. Since the source domain dataset is used to train the CNNs in the fitness evaluation, the smaller dataset will save the computational cost. The similarity between the source domain and the target domain would affect the quality of transfer learning, i.e. the performance on the target domain of the learned CNN blocks from the source domain. During the evolutionary process, each individual in the population is evaluated by the fitness evaluation at each generation, so there are a lot of fitness evaluations. Other than that, the Stochastic Gradient Descent (SGD) \cite{sutskever2013importance} algorithm used to train the CNNs in the fitness evaluation consumes most of the computational resource. Therefore, learning optimal CNN blocks in the source domain can significantly reduce the computational cost and also achieve good performance if the source domain is properly selected. 

In the target domain learning, grid search (GS) is adopted to find the final CNN that performs well on the target dataset. The stacking process generates a few CNNs based on the CNN block(s) learned from the source domain learning. The CNNs are trained by SGD and evaluated on the target domain dataset. Based on the empirical study from WideResNet \cite{zagoruyko2016wide} and NASNet \cite{zoph2018learning},  only a handful of CNNs need to be stacked and evaluated, so GS is used to search for the CNN that achieves the best classification accuracy, which will be the final CNN. Another reason for using GS is that it is easy to parallelise, so all of the CNNs can be evaluated at the same time. 

An important decision of choosing the CNN training algorithm needs to be made for the two evaluations --- the fitness evaluation in the source domain learning and the evaluation of the stacked CNNs in the target domain learning. In the proposed framework, an adaptive SGD algorithm called Adam optimisation \cite{kingma2014adam} is used as the SGD algorithm to train CNNs. The major reason is that the learning process of Adam optimisation is adjusted for different CNNs to accomplish the training task. While, for the standard SGD with a fixed learning rate or scheduled learning rates, the learning rates work well only on a set of CNNs, i.e. specific CNNs may require specific learning rates. Therefore, the Adam optimisation can perform a fairer comparison of CNNs \cite{wang2019evolving} \cite{wang2019particle}. This meets the demand of the two evaluations in the proposed EEDL framework, which are to select better CNN architectures by comparing their classification accuracies.

\subsection{Encoding Strategy}\label{SSS:generalnet_encoding}

\begin{figure}[ht]
	\centering
	\includegraphics[width=\linewidth]{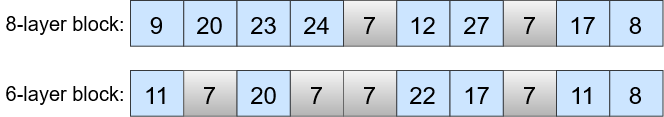}
	\caption{Two example encodings of the dense blocks.}
	\label{fig:generalnet-encoding}
\end{figure}

A similar encoding strategy as EffPNet \cite{wang2020surrogate} is proposed. 
All of the growth rates are encoded in a vector as shown in Fig. \ref{fig:generalnet-encoding}. A special value to disable a layer is set to 7
. A hyper-parameter --- the \textit{maximum number of layers} in a block need to be defined based on the hardware resource and the complexity of the dataset, which is the length of each particle. In the two examples of Fig. \ref{fig:generalnet-encoding}, the \textit{maximum number of layers} is set to 10. The top vector represents a block of 8 layers with 2 disabled layers in a grey colour. While the bottom vector represents a 6-layer block with 4 disabled layers. 

\subsection{Fitness Evaluation in Source Domain Learning} \label{SSS:generanet_fitness_evaluation}

\begin{algorithm}[h]
	\caption{Fitness Evaluation}
	\label{alg:generalnet_fitness}
	\begin{algorithmic}[1]
		\renewcommand{\algorithmicrequire}{\textbf{Input:}}
		\renewcommand{\algorithmicensure}{\textbf{Output:}}
		\newcommand{\algorithmicbreak}{\textbf{break}}
		\newcommand{\BREAK}{\STATE \algorithmicbreak}
		\REQUIRE particle $p$, \textit{source domain dataset} $d$;
		\STATE $d_{train}, d_{test}\leftarrow$ Randomly split $d$ into 80\% as the training part $d_{train}$ and 20\% as the test part $d_{test}$;
		\STATE $b\leftarrow$ Decode the particle $p$;
		\STATE $acc_{best}\leftarrow$ Use Adam optimisation \cite{kingma2014adam} to train $b$ on $d_{train}$ and evaluate $b$ on $d_{test}$ for 50 epochs, and record the best accuracy on $d_{test}$;
		\RETURN $acc_{best}$;
	\end{algorithmic}
\end{algorithm}

The fitness evaluation is similar to EffPNet \cite{wang2020surrogate}, where Adam optimisation is used to train the CNNs represented by the particles and the classification accuracy is used as the fitness value as described in Algorithm \ref{alg:generalnet_fitness}. There are a couple of points that are specific to the source domain dataset. The whole dataset in the source domain is used in the fitness evaluation, which is divided into the training part and test part by 80\% and 20\%, respectively. There is a hyper-parameter --- the \textit{total number of epochs} to train the CNNs, which is supposed to be smaller than that of EffPNet because it would be very likely to train CNNs on a much smaller dataset in the source domain with a smaller number of epochs. 

\begin{figure}[ht]
	\centering
	\includegraphics[width=\linewidth]{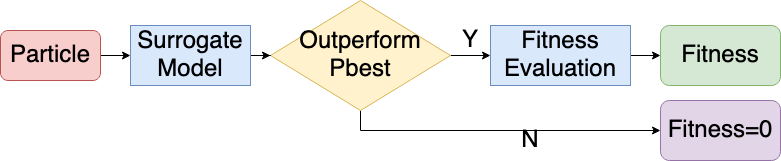}
	\caption{Surrogate-assisted fitness evaluation.}
	\label{fig:generalnet-surrogate-fitness}
\end{figure}

The surrogate model proposed in EffPNet \cite{wang2020surrogate} is integrated into the fitness evaluation to accelerate the evolutionary process as drawn in Fig. \ref{fig:generalnet-surrogate-fitness}. Before the particle is passed to the fitness evaluation, the surrogate model predicts whether the particle's current position would outperform the best-known position of the particle. If the current position is predicted to be better, the fitness evaluation in Algorithm \ref{alg:generalnet_fitness} will be used to evaluate the particle. Otherwise, the fitness value of the particle is directly set to 0. Hence, the surrogate model can filter out the evaluations of underperformed positions of the particles to save the computational cost. 

\subsection{Multi-source Knowledge Transfer with Weighted Voting}

\begin{figure*}[ht]
	\centering
	\includegraphics[width=\linewidth]{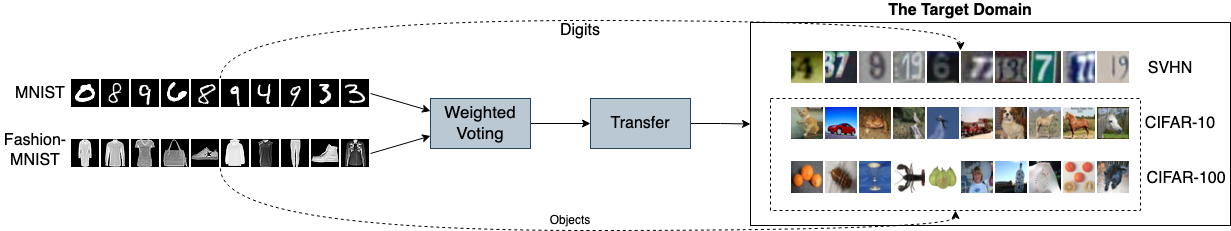}
	\caption{An example to show muli-source knowledge transfer. In the example, MNIST and fashion-MNIST are the multi-source domains. The knowledge is transferred to the target domain, where the learned block can be applied to a target domain, e.g. SVHN, CIFAR-10 or CIFAR-100.}
	\label{fig:generalnet-knowledge-transfer}
\end{figure*}

Fig. \ref{fig:generalnet-knowledge-transfer} illustrates the flow of the knowledge transferred from the multiple source domains to different target domains. When evaluating an individual/particle of the population, the fitness values on the MNIST dataset and the fashion-MNIST dataset are added together. In the experiments, the fitness values from the source domain datasets are given the same weight, so the added fitness value is then divided by 2 to get the weighted fitness value. By using the weighted voting, the knowledge learned from both the source domain datasets are passed to the target domain. The dotted lines with arrows show the directions of the knowledge transfer. The knowledge of solving image classification problems with digit pictures is gained from the MNIST dataset, and the knowledge of tackling image classification tasks with object images is obtained from the fashion-MNIST dataset. The weighted-voting mechanism combines the knowledge learned from both of the datasets together, which could be used to solve the target domain problem with either digits and objects. Therefore, the knowledge can be transferred to different target domains, e.g. the SVHN dataset with digit images, and the CIFAR datasets (including the CIFAR-10 and the CIFAR-100) containing of objects. 

\subsection{Evolving CNN Blocks in Source Domain Learning}

\begin{algorithm}[h]
	\caption{Evolving dense blocks in Source Domain}
	\label{alg:generalnet_pso}
	\begin{algorithmic}[1]
		\renewcommand{\algorithmicrequire}{\textbf{Input:}}
		\renewcommand{\algorithmicensure}{\textbf{Output:}}
		\newcommand{\algorithmicbreak}{\textbf{break}}
		\newcommand{\BREAK}{\STATE \algorithmicbreak}
		\REQUIRE generations $g$;
		\STATE $p\leftarrow$ Random initialise the particles;
		\STATE $g_{best}, j\leftarrow$ Empty, 0;
		\WHILE{$j<g$}
		\FOR{particle $part$ in $pop$}
		\STATE $i\leftarrow$ Apply PSO operations to update the position of $part$;
		\STATE $fit\leftarrow$ Use the surrogate-assisted fitness evaluation described in Section \ref{SSS:generanet_fitness_evaluation} to evaluate $part$;
		\STATE Update the fitness of $part$ by $fit$;
		\IF{$fit>$ \textit{the fitness of the particle's best}}
		\STATE Update the particle's best with $part$;
		\ENDIF
		\ENDFOR
		\STATE $g_{best}\leftarrow$ Update with the best particle among the current $g_{best}$ and $pop$;
		\STATE $j\leftarrow$ $j+1$;
		\ENDWHILE
		\RETURN $g_{best}$;
	\end{algorithmic}
\end{algorithm}

The process of evolving CNN blocks is straightforward after designing the encoding strategy and the fitness evaluation method. The pseudo-code is illustrated in Algorithm \ref{alg:generalnet_pso}. The population is initialised by randomly generating vectors based on the encoding strategy first. The standard PSO is used to update the particle positions of the population. During the evolutionary process, the global best and the particle's best are updated based on the surrogate-assisted fitness evaluation method. In the end, the global best is returned as the output solution representing the evolved block learned from the source domain. 

\subsection{Transferring Evolved Blocks to Target Domain} \label{SSS:generalnet_stacking}

\begin{figure}[ht]
	\centering
	\includegraphics[width=0.6\linewidth]{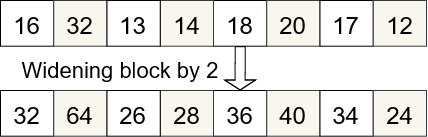}
	\caption{Widening the evolved block by 2.}
	\label{fig:generalnet-block-widending}
\end{figure}

\begin{figure}[ht]
	\centering
	\includegraphics[width=\linewidth]{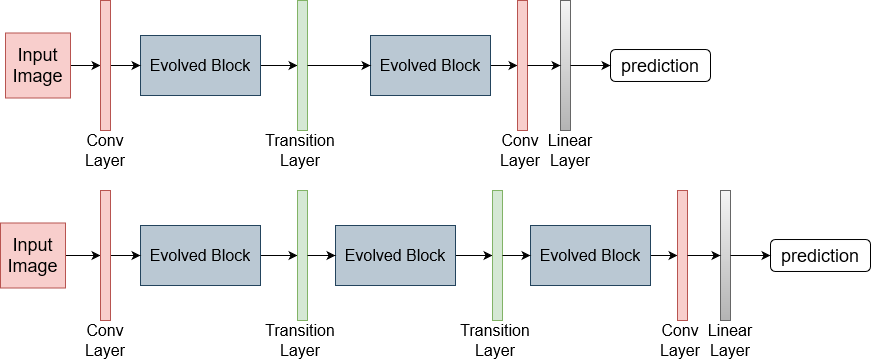}
	\caption{Deepening the evolved block by 2 and 3.}
	\label{fig:generalnet-block-deepening}
\end{figure}

Since the block is evolved from a smaller dataset, the capacity of the CNN comprised of one block might not fit the more complex target dataset. The main technique to transfer the evolved block is to stack the block to increase the capacity of CNNs to fit the target domain. The stacking method is split into two directions --- \textit{widening the CNN} and \textit{deepening the CNN}. Widening the CNN to increase the capacity is inspired by WRN \cite{zagoruyko2016wide}, which widens the feature maps of the ResNet blocks \cite{he2016deep} to achieve improved performance. Fig. \ref{fig:generalnet-block-widending} illustrates an example of the widening process of a CNN block, where the value in each dimension of the block below is obtained by the corresponding value in the block above multiplied by 2. The number of times by which the value is multiplied is a hyper-parameter called the \textit{widening factor} of the stacking process. 

The method of deepening the block is inspired by NASNet \cite{zoph2018learning}, which stacks the blocks by duplicating the same block multiple times and connecting them using a feed-forward fashion to achieve improved performance. Two stacking examples are drawn in Fig. \ref{fig:generalnet-block-deepening}. The top is an example of stacking the block twice, and the bottom example stacks the block three times. The blocks are connected by a transition layer, which is the same as the transition layer in DenseNet \cite{huang2017densely}. The other layers --- two convolutional layers and linear layers are the same as those in DenseNet as well. The number of blocks stacked in the CNN is another hyper-parameter called the \textit{deepening factor}. 

The grid search is adopted to optimise the two hyper-parameters. Based on the hardware resource and the dataset complexity, the \textit{widending factor} and the \textit{deepening factor} can be selected, but there could be only a handful of options for each of them, so the search space would be very small, which is why grid search is chosen for the optimisation. 

\section{Experimental Design}

\subsection{Benchmark Datasets}

\begin{table}[t]
	\renewcommand{\arraystretch}{1.3}
	\caption{Benchmark datasets}
	\label{table:generalnet_datasets}
	\centering
	\begin{tabular}{|p{.1\linewidth}|p{.15\linewidth}|p{.1\linewidth}|p{.1\linewidth}|p{.15\linewidth}|p{.15\linewidth}|}
		\hline
		Dataset Name & \textbf{Colour} & \textbf{Image Size} & Number of Classes & Number of Training Images & Number of Test Images  \\
		\hline
		\multicolumn{6}{|c|}{\textbf{Source Domain Datasets}} \\
		\hline
		MNIST \cite{lecun1998gradient} & greyscale & $28\times28$ & 10 & 60,000 & 10,000 \\
		\hline
		Fashion-MNIST \cite{xiao2017fashion} & greyscale & $28\times28$ & 10 & 60,000 & 10,000 \\
		\hline
		\multicolumn{6}{|c|}{\textbf{Target Domain Datasets}} \\
		\hline
		CIFAR-10 \cite{krizhevsky2009learning} & RGB colour & $32\times32$ & 10 & 50,000 & 10,000 \\
		\hline
		CIFAR-100 \cite{krizhevsky2009learning} & RGB colour & $32\times32$ & 100 & 50,000 & 10,000 \\
		\hline
		SVHN \cite{netzer2011reading} & RGB colour & $32\times32$ & 100 & 73,257 (+531,131) & 26,032 \\
		\hline
	\end{tabular}
\end{table}

Three widely-used benchmark datasets for image classification tasks --- CIFAR-10, CIFAR-100, SVHN are to evaluate the proposed method. These three datasets are the target domains. Based on the availability and similarity of smaller datasets to the target domains, the MNIST is chosen as the source domain dataset for learning the knowledge to solve image classification problem with digit images, which could be applied to the SVHN dataset. Fashion-MNIST is picked as the source domain for gaining the knowledge of tackling image classification tasks with object images, which could be transferred to the CIFAR-10 and CIFAR-100 datasets. However, Fashion-MNIST is comprised of fashion objects such as sandals and dresses, while CIFAR-10 and CIFAR-100 are composed of natural objects such as cats, dogs and air planes. The similarity between the source domain of Fashion-MNIST and the target domains of CIFAR-10 and CIFAR-100 are not very close, which could be a challenge for transfer learning methods. 

The details of the benchmark datasets in the source domain and target domain are listed in Table \ref{table:generalnet_datasets}. The source domain is considered smaller than the target domain in three ways. Firstly, the source domain datasets are greyscaled, while the target domain datasets are comprised of coloured images with three channels --- RGB. Secondly, the resolution of images in the source domain is smaller than those in the target domain. Finally, the complexity of source domain datasets is less than that of the target domain datasets, so it takes much fewer epochs to train CNNs in the source domain. Therefore, the computational cost of training CNNs on the source domain is much less than that on the target domain, which can accelerate the automated process of neural architecture search for solving the target domain problem. 

\subsection{Peer Competitors}

As CIFAR-10 is a widely-used benchmark dataset to evaluate in all three areas --- manually designed CNNs, automatically designed CNNs by reinforcement learning and automatically designed CNNs by evolutionary computation. There are many available performance results reported, so more peer competitors are taken for CIFAR-10 and CIFAR-10 is the main benchmark dataset to evaluate the proposed method. The following CNNs are chosen as the peer competitors for CIFAR-10. Two the state-of-the-art CNNs designed manually are chosen, which are ResNet \cite{he2016deep} and DenseNet \cite{huang2017densely}. The CNNs  automatically designed by reinforcement learning are  BockQNN \cite{zhong2018practical}, EAS \cite{cai2018efficient}, NASNet-A (7 @ 2304) \cite{zoph2018learning}, NASH (ensemble across runs) \cite{elsken2017simple}, NAS v3 max pooling \cite{zoph2016neural}. The CNNs automatically designed by EC methods are EIGEN \cite{ren2019eigen}, AmoebaNet-B (6,128) \cite{real2018regularized}, Hier. repr-n, evolution (7000 samples), CGP-CNN(ResSet) \cite{suganuma2017genetic}, DENSER \cite{assunccao2018evolving} GeNet from WRN \cite{xie2017genetic}, CoDeapNEAT \cite{miikkulainen2019evolving}, LS-Evolution \cite{real2017large}. 

For CIFAR-100 and SVHN, there are less results reported on these two datasets, so these two benchmark datasets are used as additional benchmark datasets to CIFAR-10 to further evaluate the proposed method. Based on the availability of the reported performance, the following peer competitors are selected, where are Network in Network \cite{lin2013network}, Deeply Supervised Net \cite{lee2015deeply}, FractalNet \cite{larsson2016fractalnet}, WideResNet \cite{zagoruyko2016wide}, ResNet \cite{huang2016deep}, DenseNet(k=12) \cite{huang2017densely}.

\subsection{Parameter Settings}

\begin{table}[t]
	\renewcommand{\arraystretch}{1.3}
	\caption{Parameter settings}
	\label{table:generalnet_parameters}
	\centering
	\begin{tabular}{|c|c|}
		\hline
		Parameter & Value\\
		\hline
		\multicolumn{2}{|c|}{\textbf{EEDL hyper-parameters}} \\
		\hline
		\textit{maximum number of layers} in Section \ref{SSS:generalnet_encoding} & 16\\
		\hline
		\textit{total number of epochs} in Section \ref{SSS:generanet_fitness_evaluation} & 50\\
		\hline
		range of \textit{widening factor} in Section \ref{SSS:generalnet_stacking} & [1--3]\\
		\hline
		range of \textit{deepening factor} in Section \ref{SSS:generalnet_stacking} & [2--5]\\
		\hline
		\multicolumn{2}{|c|}{\textbf{PSO parameters}} \\
		\hline
		inertia weight $w$ & 0.7298\\
		\hline
		acceleration coefficient $c1$ & 1.49618\\
		\hline
		acceleration coefficient $c2$ & 1.49618\\
		\hline
		velocity range & [-12.5, 12.5]\\
		\hline
		population size & 30\\
		\hline
		number of generations & 50\\
		\hline
	\end{tabular}
\end{table}

The parameters of the experiments are depicted in Table \ref{table:generalnet_parameters}. There are two parts of the parameters --- the hyper-parameters specific to the proposed method and the parameters for PSO. In the first part, the \textit{maximum number of layers} is set to 16 by considering the hardware resource and complexity of the source domain datasets. 50 is used as the \textit{total number of epochs} based on the required epochs to train CNNs on the source domain dataset. The \textit{widening factor} and the \textit{deepening factor} are designed based on the complexity of the target domain datasets and the hardware complexity. In the second part, the PSO parameters are set based on the community convention \cite{shi1998parameter} \cite{trelea2003particle} \cite{van2006study} \cite{wang2018evolving}.10 runs are performed to achieve the experimental results for statistical tests. 

\section{Result Analysis}

\subsection{Performance Comparisons on CIFAR-10}

\begin{table}[ht]
	\renewcommand{\arraystretch}{1.3}
	\small
	\caption{Performance comparison with peer competitors on CIFAR-10}
	\label{table:generalnet_performance}
	\centering
	\begin{tabular}{|P{0.2\linewidth}|P{0.2\linewidth}|P{0.2\linewidth}|P{0.2\linewidth}|}
		\hline
		Method & CIFAR-10 (Error rate\%) & Number of Parameters & Computational Cost\\
		\hline
		ResNet-110 \cite{he2016deep} & 6.43 & \textbf{1.7M} & --\\
		\hline
		DenseNet(k = 40) \cite{huang2017densely} & 3.74 & 27.2M & --\\
		\hline
		\toprule
		\hline
		BockQNN \cite{zhong2018practical} & \textit{3.54} & 39.8M & 96 GPU-days\\
		\hline
		EAS \cite{cai2018efficient} & 4.23 & 23.4M & $<$10 GPU-days\\
		\hline
		NASNet-A (7 @ 2304) \cite{zoph2018learning} & \textbf{2.97} & 27.6M & 2,000 GPU-days\\
		\hline
		NASH (ensemble across runs) \cite{elsken2017simple} & 4.40 & 88M & 4 GPU-days\\
		\hline
		NAS v3 max pooling \cite{zoph2016neural} & 4.47 & 7.1M & 22,400 GPU-days\\
		\hline
		\toprule
		\hline
		EIGEN \cite{ren2019eigen} & 5.4 & \textit{2.6M} & 2 GPU-days\\
		\hline
		RENAS \cite{chen2019renas} & \textbf{2.88} & 3.5M & 6 GPU-days\\
		\hline
		AmoebaNet-B (6,128) \cite{real2018regularized} & \textbf{2.98} & 34.9M & 3150 GPU-days\\
		\hline
		Hier. repr-n, evolution (7000 samples) \cite{liu2017hierarchical} & 3.75 & -- & 300 GPU-days\\
		\hline
		HGAPSO \cite{wang2019hybrid} & 4.37 & -- & 7 GPU-days\\
		\hline
		CGP-CNN(ResSet) \cite{suganuma2017genetic} & 5.98 & \textbf{1.68M} & 29.8 GPU-days\\
		\hline
		DENSER \cite{assunccao2018evolving} & 5.87 & 10.81M & --\\
		\hline 
		GeNet from WRN \cite{xie2017genetic} & 5.39 & -- & 100 GPU-days\\
		\hline
		CoDeapNEAT \cite{miikkulainen2019evolving} & 7.3 & -- & --\\
		\hline
		LS-Evolution \cite{real2017large} & 4.4 & 40.4M & $>$2,730 GPU-days\\
		\hline
		\toprule
		\hline
		\textbf{EEDL (Best classification accuracy)} & 3.46 & 2.29M & $<$ 40 GPU-hours\\
		\hline
		\textbf{EEDL (10 runs)} & 3.53$\pm$0.0092 & 2.41M$\pm$0.04M & $<$ 40 GPU-hours\\
		\hline
	\end{tabular}
\end{table}

Table \ref{table:generalnet_performance} lists the performance on the CIFAR-10 dataset including the error rate, the number of parameters and the computational cost of searching the network architectures (including both the source and target domain learning). The bold values mean the corresponding competitors outperform the proposed method. While the others indicate the proposed method excels. For the proposed method, the results from 10 runs are presented at the bottom of the table with \textit{the best value} and \textit{the mean value} $\pm$ \textit{the standard deviation}. In regard to the error rate, i.e. the classification accuracy, three out of the 17 peer competitors achieve better performance than the proposed method. While BlockQNN \citep{zhong2018practical} seems competitive to the proposed method, but by applying Mann-Whitney-Wilcoxon (MWW) statistical test, the statistical difference is still significant, so the proposed method keeps the 4th place among the 18 methods shown in the table. However, the CNN models from three competitors with smaller error rates substantially outsize the models obtained by the proposed method. In terms of the number of parameters, the proposed method ranks the 3rd among the 17 peer competitors and itself. Again, the two competitors with smaller model sizes significantly sacrifice the classification comparing to the proposed method. Looking at the computational cost, the proposed method takes only 40 GPU-hours to accomplish the neural architecture search task, which is faster than all of the 17 peer competitors. Overall, the proposed has demonstrated the superior performance against the 17 peer competitors on the CIFAR-10 dataset by considering the classification accuracy, the model size and the computational cost. 

\subsection{Performance Comparison on CIFAR-100 and SVHN}

\begin{table}[ht]
	\renewcommand{\arraystretch}{1.3}
	\caption{Error rate comparison with peer competitors on CIFAR-100 and SVHN}
	\label{table:generalnet_performance_tl}
	\centering
	\begin{tabular}{|P{0.4\linewidth}|P{0.2\linewidth}|P{0.2\linewidth}|}
		\hline
		Method & CIFAR-100 & SVHN\\
		\hline
		\hline
		Network in Network \cite{lin2013network} & 35.68 & 2.35 \\
		\hline
		Deeply Supervised Net \cite{lee2015deeply} & 34.57 & 1.92 \\
		\hline
		FractalNet \cite{larsson2016fractalnet} & 23.30 & 2.01 \\
		\hline
		Wide ResNet \cite{zagoruyko2016wide} & 22.07 & \textit{1.85} \\
		\hline
		ResNet \cite{huang2016deep} & 27.22 & 2.01 \\
		\hline
		DenseNet(k=12) \cite{huang2017densely} & 20.20 & \textbf{1.67} \\
		\hline
		\hline
		\textbf{EEDL (Best)} & 18.36 &  1.76 \\
		\hline
		\textbf{EEDL (10 runs)} & 18.90$\pm$0.2351 & 1.78$\pm$0.0196 \\
		\hline
	\end{tabular}
\end{table}

To further evaluate the proposed method, especially assessing its transferability and generalisation, the error rates on the CIFAR100 and SVHN datasets of 6 peer competitors and the proposed method are exhibited in Table \ref{table:generalnet_performance_tl}. On the CIFAR-100 dataset, the proposed method achieves the best classification accuracy. While on the SVHN dataset, only one of the 6 peer competitors outperforms the proposed method with a tiny margin. The evolutionary process of the neural architecture search only performs once on the source domain datasets to produce a transferable block, which then is stacked based on the target domain. Since the proposed method has demonstrated its strong competitiveness on all three target domain datasets, it demonstrates the strong transferability of the evolved transferable block and the good generalisation of the proposed method. 






\section{Conclusions and Future Work}

The overall goal of proposing a new evolutionary deep learning framework has been successfully achieved. From the experimental results, the proposed method has achieved very competitive performance in terms of the classification accuracy, the model size of the evolved CNNs, and the computational cost.  To be more specific with the contributions, the proposed framework is decoupled into a two-stage learning process --- the source domain learning and the target domain learning. In the source domain learning, a PSO-based method is proposed to efficiently and effectively evolve a CNN block from the smaller datasets in the source domains. In the target domain learning, both the widening and deepening methods are proposed to stack the evolved CNN block(s) to enlarge the capacity of CNN models.  Then, a grid search based method is proposed to find good CNN architectures with larger capacities to keep the classification accuracy of the proposed method competitive with the state-of-the-art methods.  

Although the proposed method has achieved very competitive performance, there is still some future work that could be done. Firstly, as the proposed EDL framework has demonstrated its efficiency and effectiveness of evolving deep CNNs on the image classification task, it would be interesting to perform experiments to evolve other neural networks, e.g. recurrent neural networks. on the text classification tasks. Secondly, the proposed EDL framework is still a black-box, which only shows the performance of the final evolved CNNs. It would be more interesting to find a way to visualise the knowledge and interpret the flow of the knowledge transfer, e.g. to analyse what knowledge is learned from the target domain and which part of the knowledge is transferred to the target domain.

\bibliographystyle{ACM-Reference-Format}
\bibliography{generalnet-gecco} 


\begin{thebibliography}{47}


\ifx \showCODEN    \undefined \def \showCODEN     #1{\unskip}     \fi
\ifx \showDOI      \undefined \def \showDOI       #1{#1}\fi
\ifx \showISBNx    \undefined \def \showISBNx     #1{\unskip}     \fi
\ifx \showISBNxiii \undefined \def \showISBNxiii  #1{\unskip}     \fi
\ifx \showISSN     \undefined \def \showISSN      #1{\unskip}     \fi
\ifx \showLCCN     \undefined \def \showLCCN      #1{\unskip}     \fi
\ifx \shownote     \undefined \def \shownote      #1{#1}          \fi
\ifx \showarticletitle \undefined \def \showarticletitle #1{#1}   \fi
\ifx \showURL      \undefined \def \showURL       {\relax}        \fi
\providecommand\bibfield[2]{#2}
\providecommand\bibinfo[2]{#2}
\providecommand\natexlab[1]{#1}
\providecommand\showeprint[2][]{arXiv:#2}

\bibitem[\protect\citeauthoryear{Assun{\c{c}}{\~a}o, Louren{\c{c}}o, Machado,
  and Ribeiro}{Assun{\c{c}}{\~a}o et~al\mbox{.}}{2018}]%
        {assunccao2018evolving}
\bibfield{author}{\bibinfo{person}{Filipe Assun{\c{c}}{\~a}o},
  \bibinfo{person}{Nuno Louren{\c{c}}o}, \bibinfo{person}{Penousal Machado},
  {and} \bibinfo{person}{Bernardete Ribeiro}.} \bibinfo{year}{2018}\natexlab{}.
\newblock \showarticletitle{Evolving the topology of large scale deep neural
  networks}. In \bibinfo{booktitle}{{\em European Conference on Genetic
  Programming}}. Springer, \bibinfo{pages}{19--34}.
\newblock


\bibitem[\protect\citeauthoryear{Brock, Lim, Ritchie, and Weston}{Brock
  et~al\mbox{.}}{2018}]%
        {brock2017smash}
\bibfield{author}{\bibinfo{person}{Andrew Brock}, \bibinfo{person}{Theodore
  Lim}, \bibinfo{person}{James~M. Ritchie}, {and} \bibinfo{person}{Nick
  Weston}.} \bibinfo{year}{2018}\natexlab{}.
\newblock \showarticletitle{{SMASH:} One-Shot Model Architecture Search through
  HyperNetworks}. In \bibinfo{booktitle}{{\em 6th International Conference on
  Learning Representations (ICLR)}}.
\newblock
\showURL{%
\url{http://arxiv.org/abs/1708.05344}}


\bibitem[\protect\citeauthoryear{Cai, Chen, Zhang, Yu, and Wang}{Cai
  et~al\mbox{.}}{2018}]%
        {cai2018efficient}
\bibfield{author}{\bibinfo{person}{Han Cai}, \bibinfo{person}{Tianyao Chen},
  \bibinfo{person}{Weinan Zhang}, \bibinfo{person}{Yong Yu}, {and}
  \bibinfo{person}{Jun Wang}.} \bibinfo{year}{2018}\natexlab{}.
\newblock \showarticletitle{Efficient architecture search by network
  transformation}. In \bibinfo{booktitle}{{\em Thirty-Second AAAI Conference on
  Artificial Intelligence}}. \bibinfo{pages}{2787--2794}.
\newblock


\bibitem[\protect\citeauthoryear{Chang and Lin}{Chang and Lin}{2011}]%
        {chang2011libsvm}
\bibfield{author}{\bibinfo{person}{Chih-Chung Chang} {and}
  \bibinfo{person}{Chih-Jen Lin}.} \bibinfo{year}{2011}\natexlab{}.
\newblock \showarticletitle{LIBSVM: A library for support vector machines}.
\newblock \bibinfo{journal}{{\em ACM transactions on intelligent systems and
  technology (TIST)\/}} \bibinfo{volume}{2}, \bibinfo{number}{3}
  (\bibinfo{year}{2011}), \bibinfo{pages}{1--27}.
\newblock


\bibitem[\protect\citeauthoryear{Chen, Meng, Zhang, Xiang, Huang, Mu, and
  Wang}{Chen et~al\mbox{.}}{2019}]%
        {chen2019renas}
\bibfield{author}{\bibinfo{person}{Yukang Chen}, \bibinfo{person}{Gaofeng
  Meng}, \bibinfo{person}{Qian Zhang}, \bibinfo{person}{Shiming Xiang},
  \bibinfo{person}{Chang Huang}, \bibinfo{person}{Lisen Mu}, {and}
  \bibinfo{person}{Xinggang Wang}.} \bibinfo{year}{2019}\natexlab{}.
\newblock \showarticletitle{RENAS: Reinforced Evolutionary Neural Architecture
  Search}. In \bibinfo{booktitle}{{\em Proceedings of the IEEE Conference on
  Computer Vision and Pattern Recognition}}. \bibinfo{pages}{4787--4796}.
\newblock


\bibitem[\protect\citeauthoryear{Cortes and Vapnik}{Cortes and Vapnik}{1995}]%
        {cortes1995support}
\bibfield{author}{\bibinfo{person}{Corinna Cortes} {and}
  \bibinfo{person}{Vladimir Vapnik}.} \bibinfo{year}{1995}\natexlab{}.
\newblock \showarticletitle{Support-vector networks}.
\newblock \bibinfo{journal}{{\em Machine learning\/}} \bibinfo{volume}{20},
  \bibinfo{number}{3} (\bibinfo{year}{1995}), \bibinfo{pages}{273--297}.
\newblock


\bibitem[\protect\citeauthoryear{Eberhart and Kennedy}{Eberhart and
  Kennedy}{1995}]%
        {eberhart1995new}
\bibfield{author}{\bibinfo{person}{Russell Eberhart} {and}
  \bibinfo{person}{James Kennedy}.} \bibinfo{year}{1995}\natexlab{}.
\newblock \showarticletitle{A new optimizer using particle swarm theory}. In
  \bibinfo{booktitle}{{\em MHS'95. Proceedings of the Sixth International
  Symposium on Micro Machine and Human Science}}. Ieee,
  \bibinfo{pages}{39--43}.
\newblock


\bibitem[\protect\citeauthoryear{Elsken, Metzen, and Hutter}{Elsken
  et~al\mbox{.}}{2018}]%
        {elsken2017simple}
\bibfield{author}{\bibinfo{person}{Thomas Elsken}, \bibinfo{person}{Jan~Hendrik
  Metzen}, {and} \bibinfo{person}{Frank Hutter}.}
  \bibinfo{year}{2018}\natexlab{}.
\newblock \showarticletitle{Simple and efficient architecture search for
  Convolutional Neural Networks}. In \bibinfo{booktitle}{{\em 6th International
  Conference on Learning Representations (ICLR)}}.
\newblock
\showURL{%
\url{https://arxiv.org/abs/1711.04528}}


\bibitem[\protect\citeauthoryear{Han, Kim, and Kim}{Han et~al\mbox{.}}{2017}]%
        {han2017deep}
\bibfield{author}{\bibinfo{person}{Dongyoon Han}, \bibinfo{person}{Jiwhan Kim},
  {and} \bibinfo{person}{Junmo Kim}.} \bibinfo{year}{2017}\natexlab{}.
\newblock \showarticletitle{Deep pyramidal residual networks}. In
  \bibinfo{booktitle}{{\em Proceedings of the IEEE Conference on Computer
  Vision and Pattern Recognition}}. \bibinfo{pages}{5927--5935}.
\newblock


\bibitem[\protect\citeauthoryear{He, Zhang, Ren, and Sun}{He
  et~al\mbox{.}}{2016}]%
        {he2016deep}
\bibfield{author}{\bibinfo{person}{Kaiming He}, \bibinfo{person}{Xiangyu
  Zhang}, \bibinfo{person}{Shaoqing Ren}, {and} \bibinfo{person}{Jian Sun}.}
  \bibinfo{year}{2016}\natexlab{}.
\newblock \showarticletitle{Deep residual learning for image recognition}. In
  \bibinfo{booktitle}{{\em Proceedings of the IEEE conference on computer
  vision and pattern recognition}}. \bibinfo{pages}{770--778}.
\newblock


\bibitem[\protect\citeauthoryear{Huang, Liu, Van Der~Maaten, and
  Weinberger}{Huang et~al\mbox{.}}{2017}]%
        {huang2017densely}
\bibfield{author}{\bibinfo{person}{Gao Huang}, \bibinfo{person}{Zhuang Liu},
  \bibinfo{person}{Laurens Van Der~Maaten}, {and} \bibinfo{person}{Kilian~Q
  Weinberger}.} \bibinfo{year}{2017}\natexlab{}.
\newblock \showarticletitle{Densely connected convolutional networks}. In
  \bibinfo{booktitle}{{\em Proceedings of the IEEE conference on computer
  vision and pattern recognition}}. \bibinfo{pages}{4700--4708}.
\newblock


\bibitem[\protect\citeauthoryear{Huang, Sun, Liu, Sedra, and Weinberger}{Huang
  et~al\mbox{.}}{2016}]%
        {huang2016deep}
\bibfield{author}{\bibinfo{person}{Gao Huang}, \bibinfo{person}{Yu Sun},
  \bibinfo{person}{Zhuang Liu}, \bibinfo{person}{Daniel Sedra}, {and}
  \bibinfo{person}{Kilian~Q Weinberger}.} \bibinfo{year}{2016}\natexlab{}.
\newblock \showarticletitle{Deep networks with stochastic depth}. In
  \bibinfo{booktitle}{{\em European conference on computer vision}}. Springer,
  \bibinfo{pages}{646--661}.
\newblock


\bibitem[\protect\citeauthoryear{Kennedy and Eberhart}{Kennedy and
  Eberhart}{1995}]%
        {kennedy1995particle}
\bibfield{author}{\bibinfo{person}{James Kennedy} {and}
  \bibinfo{person}{Russell Eberhart}.} \bibinfo{year}{1995}\natexlab{}.
\newblock \showarticletitle{Particle swarm optimization}. In
  \bibinfo{booktitle}{{\em Proceedings of ICNN'95-International Conference on
  Neural Networks}}, Vol.~\bibinfo{volume}{4}. IEEE,
  \bibinfo{pages}{1942--1948}.
\newblock


\bibitem[\protect\citeauthoryear{Kingma and Ba}{Kingma and Ba}{2015}]%
        {kingma2014adam}
\bibfield{author}{\bibinfo{person}{Diederik~P. Kingma} {and}
  \bibinfo{person}{Jimmy Ba}.} \bibinfo{year}{2015}\natexlab{}.
\newblock \showarticletitle{Adam: {A} Method for Stochastic Optimization}. In
  \bibinfo{booktitle}{{\em 3rd International Conference on Learning
  Representations (ICLR)}}.
\newblock
\showURL{%
\url{http://arxiv.org/abs/1412.6980}}


\bibitem[\protect\citeauthoryear{Krizhevsky and Hinton}{Krizhevsky and
  Hinton}{2009}]%
        {krizhevsky2009learning}
\bibfield{author}{\bibinfo{person}{Alex Krizhevsky} {and}
  \bibinfo{person}{Geoffrey Hinton}.} \bibinfo{year}{2009}\natexlab{}.
\newblock \bibinfo{booktitle}{{\em Learning multiple layers of features from
  tiny images}}.
\newblock \bibinfo{type}{{T}echnical {R}eport}.
  \bibinfo{institution}{Citeseer}.
\newblock


\bibitem[\protect\citeauthoryear{Krizhevsky, Sutskever, and Hinton}{Krizhevsky
  et~al\mbox{.}}{2012}]%
        {krizhevsky2012imagenet}
\bibfield{author}{\bibinfo{person}{Alex Krizhevsky}, \bibinfo{person}{Ilya
  Sutskever}, {and} \bibinfo{person}{Geoffrey~E Hinton}.}
  \bibinfo{year}{2012}\natexlab{}.
\newblock \showarticletitle{Imagenet classification with deep convolutional
  neural networks}. In \bibinfo{booktitle}{{\em Advances in neural information
  processing systems}}. \bibinfo{pages}{1097--1105}.
\newblock


\bibitem[\protect\citeauthoryear{Larsson, Maire, and Shakhnarovich}{Larsson
  et~al\mbox{.}}{2017}]%
        {larsson2016fractalnet}
\bibfield{author}{\bibinfo{person}{Gustav Larsson}, \bibinfo{person}{Michael
  Maire}, {and} \bibinfo{person}{Gregory Shakhnarovich}.}
  \bibinfo{year}{2017}\natexlab{}.
\newblock \showarticletitle{FractalNet: Ultra-Deep Neural Networks without
  Residuals}. In \bibinfo{booktitle}{{\em 5th International Conference on
  Learning Representations (ICLR)}}.
\newblock
\showURL{%
\url{https://arxiv.org/abs/1605.07648}}


\bibitem[\protect\citeauthoryear{LeCun, Bottou, Bengio, and Haffner}{LeCun
  et~al\mbox{.}}{1998}]%
        {lecun1998gradient}
\bibfield{author}{\bibinfo{person}{Yann LeCun}, \bibinfo{person}{L{\'e}on
  Bottou}, \bibinfo{person}{Yoshua Bengio}, {and} \bibinfo{person}{Patrick
  Haffner}.} \bibinfo{year}{1998}\natexlab{}.
\newblock \showarticletitle{Gradient-based learning applied to document
  recognition}.
\newblock \bibinfo{journal}{{\it Proc. IEEE}} \bibinfo{volume}{86},
  \bibinfo{number}{11} (\bibinfo{year}{1998}), \bibinfo{pages}{2278--2324}.
\newblock


\bibitem[\protect\citeauthoryear{Lee, Xie, Gallagher, Zhang, and Tu}{Lee
  et~al\mbox{.}}{2015}]%
        {lee2015deeply}
\bibfield{author}{\bibinfo{person}{Chen-Yu Lee}, \bibinfo{person}{Saining Xie},
  \bibinfo{person}{Patrick Gallagher}, \bibinfo{person}{Zhengyou Zhang}, {and}
  \bibinfo{person}{Zhuowen Tu}.} \bibinfo{year}{2015}\natexlab{}.
\newblock \showarticletitle{Deeply-supervised nets}. In
  \bibinfo{booktitle}{{\em Artificial intelligence and statistics}}.
  \bibinfo{pages}{562--570}.
\newblock


\bibitem[\protect\citeauthoryear{Lin, Chen, and Yan}{Lin et~al\mbox{.}}{2014}]%
        {lin2013network}
\bibfield{author}{\bibinfo{person}{Min Lin}, \bibinfo{person}{Qiang Chen},
  {and} \bibinfo{person}{Shuicheng Yan}.} \bibinfo{year}{2014}\natexlab{}.
\newblock \showarticletitle{Network In Network}. In \bibinfo{booktitle}{{\em
  2nd International Conference on Learning Representations (ICLR)}}.
\newblock
\showURL{%
\url{http://arxiv.org/abs/1312.4400}}


\bibitem[\protect\citeauthoryear{Liu, Simonyan, Vinyals, Fernando, and
  Kavukcuoglu}{Liu et~al\mbox{.}}{2018}]%
        {liu2017hierarchical}
\bibfield{author}{\bibinfo{person}{Hanxiao Liu}, \bibinfo{person}{Karen
  Simonyan}, \bibinfo{person}{Oriol Vinyals}, \bibinfo{person}{Chrisantha
  Fernando}, {and} \bibinfo{person}{Koray Kavukcuoglu}.}
  \bibinfo{year}{2018}\natexlab{}.
\newblock \showarticletitle{Hierarchical Representations for Efficient
  Architecture Search}. In \bibinfo{booktitle}{{\em 6th International
  Conference on Learning Representations (ICLR)}}.
\newblock
\showURL{%
\url{https://arxiv.org/abs/1711.00436}}


\bibitem[\protect\citeauthoryear{Lu, Deb, Goodman, Banzhaf, and Boddeti}{Lu
  et~al\mbox{.}}{2020}]%
        {lu2020nsganetv2}
\bibfield{author}{\bibinfo{person}{Zhichao Lu}, \bibinfo{person}{Kalyanmoy
  Deb}, \bibinfo{person}{Erik Goodman}, \bibinfo{person}{Wolfgang Banzhaf},
  {and} \bibinfo{person}{Vishnu~Naresh Boddeti}.}
  \bibinfo{year}{2020}\natexlab{}.
\newblock \showarticletitle{Nsganetv2: Evolutionary multi-objective
  surrogate-assisted neural architecture search}. In \bibinfo{booktitle}{{\em
  European Conference on Computer Vision}}. Springer, \bibinfo{pages}{35--51}.
\newblock


\bibitem[\protect\citeauthoryear{Miikkulainen, Liang, Meyerson, Rawal, Fink,
  Francon, Raju, Shahrzad, Navruzyan, Duffy, et~al\mbox{.}}{Miikkulainen
  et~al\mbox{.}}{2019}]%
        {miikkulainen2019evolving}
\bibfield{author}{\bibinfo{person}{Risto Miikkulainen}, \bibinfo{person}{Jason
  Liang}, \bibinfo{person}{Elliot Meyerson}, \bibinfo{person}{Aditya Rawal},
  \bibinfo{person}{Daniel Fink}, \bibinfo{person}{Olivier Francon},
  \bibinfo{person}{Bala Raju}, \bibinfo{person}{Hormoz Shahrzad},
  \bibinfo{person}{Arshak Navruzyan}, \bibinfo{person}{Nigel Duffy},
  {et~al\mbox{.}}} \bibinfo{year}{2019}\natexlab{}.
\newblock \showarticletitle{Evolving deep neural networks}.
\newblock In \bibinfo{booktitle}{{\em Artificial Intelligence in the Age of
  Neural Networks and Brain Computing}}. \bibinfo{publisher}{Elsevier},
  \bibinfo{pages}{293--312}.
\newblock


\bibitem[\protect\citeauthoryear{Netzer, Wang, Coates, Bissacco, Wu, and
  Ng}{Netzer et~al\mbox{.}}{2011}]%
        {netzer2011reading}
\bibfield{author}{\bibinfo{person}{Yuval Netzer}, \bibinfo{person}{Tao Wang},
  \bibinfo{person}{Adam Coates}, \bibinfo{person}{Alessandro Bissacco},
  \bibinfo{person}{Bo Wu}, {and} \bibinfo{person}{Andrew~Y Ng}.}
  \bibinfo{year}{2011}\natexlab{}.
\newblock \showarticletitle{Reading digits in natural images with unsupervised
  feature learning}.
\newblock  (\bibinfo{year}{2011}).
\newblock


\bibitem[\protect\citeauthoryear{Real, Aggarwal, Huang, and Le}{Real
  et~al\mbox{.}}{2019}]%
        {real2018regularized}
\bibfield{author}{\bibinfo{person}{Esteban Real}, \bibinfo{person}{Alok
  Aggarwal}, \bibinfo{person}{Yanping Huang}, {and} \bibinfo{person}{Quoc~V.
  Le}.} \bibinfo{year}{2019}\natexlab{}.
\newblock \showarticletitle{Regularized Evolution for Image Classifier
  Architecture Search}. In \bibinfo{booktitle}{{\em The Thirty-Third {AAAI}
  Conference on Artificial Intelligence (AAAI)}}. \bibinfo{pages}{4780--4789}.
\newblock


\bibitem[\protect\citeauthoryear{Real, Moore, Selle, Saxena, Suematsu, Tan, Le,
  and Kurakin}{Real et~al\mbox{.}}{2017}]%
        {real2017large}
\bibfield{author}{\bibinfo{person}{Esteban Real}, \bibinfo{person}{Sherry
  Moore}, \bibinfo{person}{Andrew Selle}, \bibinfo{person}{Saurabh Saxena},
  \bibinfo{person}{Yutaka~Leon Suematsu}, \bibinfo{person}{Jie Tan},
  \bibinfo{person}{Quoc~V Le}, {and} \bibinfo{person}{Alexey Kurakin}.}
  \bibinfo{year}{2017}\natexlab{}.
\newblock \showarticletitle{Large-scale evolution of image classifiers}. In
  \bibinfo{booktitle}{{\em Proceedings of the 34th International Conference on
  Machine Learning-Volume 70}}. JMLR. org, \bibinfo{pages}{2902--2911}.
\newblock


\bibitem[\protect\citeauthoryear{Ren, Li, Yang, Xu, Yang, and Foran}{Ren
  et~al\mbox{.}}{2019}]%
        {ren2019eigen}
\bibfield{author}{\bibinfo{person}{Jian Ren}, \bibinfo{person}{Zhe Li},
  \bibinfo{person}{Jianchao Yang}, \bibinfo{person}{Ning Xu},
  \bibinfo{person}{Tianbao Yang}, {and} \bibinfo{person}{David~J Foran}.}
  \bibinfo{year}{2019}\natexlab{}.
\newblock \showarticletitle{EIGEN: Ecologically-Inspired GENetic Approach for
  Neural Network Structure Searching From Scratch}. In \bibinfo{booktitle}{{\em
  Proceedings of the IEEE Conference on Computer Vision and Pattern
  Recognition}}. \bibinfo{pages}{9059--9068}.
\newblock


\bibitem[\protect\citeauthoryear{Shi and Eberhart}{Shi and Eberhart}{1998a}]%
        {shi1998modified}
\bibfield{author}{\bibinfo{person}{Yuhui Shi} {and} \bibinfo{person}{Russell
  Eberhart}.} \bibinfo{year}{1998}\natexlab{a}.
\newblock \showarticletitle{A modified particle swarm optimizer}. In
  \bibinfo{booktitle}{{\em 1998 IEEE international conference on evolutionary
  computation proceedings. IEEE world congress on computational intelligence
  (Cat. No. 98TH8360)}}. \bibinfo{pages}{69--73}.
\newblock


\bibitem[\protect\citeauthoryear{Shi and Eberhart}{Shi and Eberhart}{1998b}]%
        {shi1998parameter}
\bibfield{author}{\bibinfo{person}{Yuhui Shi} {and} \bibinfo{person}{Russell~C
  Eberhart}.} \bibinfo{year}{1998}\natexlab{b}.
\newblock \showarticletitle{Parameter selection in particle swarm
  optimization}. In \bibinfo{booktitle}{{\em International conference on
  evolutionary programming}}. Springer, \bibinfo{pages}{591--600}.
\newblock


\bibitem[\protect\citeauthoryear{Simonyan and Zisserman}{Simonyan and
  Zisserman}{2015}]%
        {simonyan2014very}
\bibfield{author}{\bibinfo{person}{Karen Simonyan} {and}
  \bibinfo{person}{Andrew Zisserman}.} \bibinfo{year}{2015}\natexlab{}.
\newblock \showarticletitle{Very Deep Convolutional Networks for Large-Scale
  Image Recognition}. In \bibinfo{booktitle}{{\em 3rd International Conference
  on Learning Representations (ICLR)}}.
\newblock
\showURL{%
\url{http://arxiv.org/abs/1409.1556}}


\bibitem[\protect\citeauthoryear{Suganuma, Shirakawa, and Nagao}{Suganuma
  et~al\mbox{.}}{2017}]%
        {suganuma2017genetic}
\bibfield{author}{\bibinfo{person}{Masanori Suganuma},
  \bibinfo{person}{Shinichi Shirakawa}, {and} \bibinfo{person}{Tomoharu
  Nagao}.} \bibinfo{year}{2017}\natexlab{}.
\newblock \showarticletitle{A genetic programming approach to designing
  convolutional neural network architectures}. In \bibinfo{booktitle}{{\em
  Proceedings of the Genetic and Evolutionary Computation Conference}}. ACM,
  \bibinfo{pages}{497--504}.
\newblock


\bibitem[\protect\citeauthoryear{Sutskever, Martens, Dahl, and
  Hinton}{Sutskever et~al\mbox{.}}{2013}]%
        {sutskever2013importance}
\bibfield{author}{\bibinfo{person}{Ilya Sutskever}, \bibinfo{person}{James
  Martens}, \bibinfo{person}{George~E Dahl}, {and} \bibinfo{person}{Geoffrey~E
  Hinton}.} \bibinfo{year}{2013}\natexlab{}.
\newblock \showarticletitle{On the importance of initialization and momentum in
  deep learning}.
\newblock \bibinfo{journal}{{\em ICML (3)\/}} \bibinfo{volume}{28},
  \bibinfo{number}{1139-1147} (\bibinfo{year}{2013}), \bibinfo{pages}{5}.
\newblock


\bibitem[\protect\citeauthoryear{Trelea}{Trelea}{2003}]%
        {trelea2003particle}
\bibfield{author}{\bibinfo{person}{Ioan~Cristian Trelea}.}
  \bibinfo{year}{2003}\natexlab{}.
\newblock \showarticletitle{The particle swarm optimization algorithm:
  convergence analysis and parameter selection}.
\newblock \bibinfo{journal}{{\em Information processing letters\/}}
  \bibinfo{volume}{85}, \bibinfo{number}{6} (\bibinfo{year}{2003}),
  \bibinfo{pages}{317--325}.
\newblock


\bibitem[\protect\citeauthoryear{Van~den Bergh and Engelbrecht}{Van~den Bergh
  and Engelbrecht}{2006}]%
        {van2006study}
\bibfield{author}{\bibinfo{person}{Frans Van~den Bergh} {and}
  \bibinfo{person}{Andries~Petrus Engelbrecht}.}
  \bibinfo{year}{2006}\natexlab{}.
\newblock \showarticletitle{A study of particle swarm optimization particle
  trajectories}.
\newblock \bibinfo{journal}{{\em Information sciences\/}}
  \bibinfo{volume}{176}, \bibinfo{number}{8} (\bibinfo{year}{2006}),
  \bibinfo{pages}{937--971}.
\newblock


\bibitem[\protect\citeauthoryear{Wang, Sun, Xue, and Zhang}{Wang
  et~al\mbox{.}}{2018a}]%
        {wang2018evolving}
\bibfield{author}{\bibinfo{person}{Bin Wang}, \bibinfo{person}{Yanan Sun},
  \bibinfo{person}{Bing Xue}, {and} \bibinfo{person}{Mengjie Zhang}.}
  \bibinfo{year}{2018}\natexlab{a}.
\newblock \showarticletitle{Evolving deep convolutional neural networks by
  variable-length particle swarm optimization for image classification}. In
  \bibinfo{booktitle}{{\em IEEE Congress on Evolutionary Computation (CEC)}}.
  \bibinfo{pages}{1--8}.
\newblock


\bibitem[\protect\citeauthoryear{Wang, Sun, Xue, and Zhang}{Wang
  et~al\mbox{.}}{2018b}]%
        {wang2018hybrid}
\bibfield{author}{\bibinfo{person}{Bin Wang}, \bibinfo{person}{Yanan Sun},
  \bibinfo{person}{Bing Xue}, {and} \bibinfo{person}{Mengjie Zhang}.}
  \bibinfo{year}{2018}\natexlab{b}.
\newblock \showarticletitle{A hybrid differential evolution approach to
  designing deep convolutional neural networks for image classification}. In
  \bibinfo{booktitle}{{\em Australasian Joint Conference on Artificial
  Intelligence}}. Springer, \bibinfo{pages}{237--250}.
\newblock


\bibitem[\protect\citeauthoryear{Wang, Sun, Xue, and Zhang}{Wang
  et~al\mbox{.}}{2019a}]%
        {wang2019evolving}
\bibfield{author}{\bibinfo{person}{Bin Wang}, \bibinfo{person}{Yanan Sun},
  \bibinfo{person}{Bing Xue}, {and} \bibinfo{person}{Mengjie Zhang}.}
  \bibinfo{year}{2019}\natexlab{a}.
\newblock \showarticletitle{Evolving Deep Neural Networks by Multi-objective
  Particle Swarm Optimization for Image Classification}. In
  \bibinfo{booktitle}{{\em Proceedings of the Genetic and Evolutionary
  Computation Conference(GECCO)}}. \bibinfo{publisher}{ACM},
  \bibinfo{pages}{490--498}.
\newblock
\showISBNx{978-1-4503-6111-8}


\bibitem[\protect\citeauthoryear{Wang, Xue, and Zhang}{Wang
  et~al\mbox{.}}{2019b}]%
        {wang2019hybrid}
\bibfield{author}{\bibinfo{person}{Bin Wang}, \bibinfo{person}{Bing Xue}, {and}
  \bibinfo{person}{Mengjie Zhang}.} \bibinfo{year}{2019}\natexlab{b}.
\newblock \showarticletitle{A Hybrid {GA-PSO} Method for Evolving Architecture
  and Short Connections of Deep Convolutional Neural Networks}. In
  \bibinfo{booktitle}{{\em PRICAI 2019: Trends in Artificial Intelligence}}.
  \bibinfo{publisher}{Springer}, \bibinfo{pages}{650--663}.
\newblock
\showISBNx{978-3-030-29894-4}


\bibitem[\protect\citeauthoryear{Wang, Xue, and Zhang}{Wang
  et~al\mbox{.}}{2019c}]%
        {wang2019particle}
\bibfield{author}{\bibinfo{person}{Bin Wang}, \bibinfo{person}{Bing Xue}, {and}
  \bibinfo{person}{Mengjie Zhang}.} \bibinfo{year}{2019}\natexlab{c}.
\newblock \showarticletitle{Particle Swarm Optimisation for Evolving Deep
  Neural Networks for Image Classification by Evolving and Stacking
  Transferable Blocks}.
\newblock \bibinfo{journal}{{\em arXiv preprint arXiv:1907.12659\/}}
  (\bibinfo{year}{2019}).
\newblock


\bibitem[\protect\citeauthoryear{Wang, Xue, and Zhang}{Wang
  et~al\mbox{.}}{2020a}]%
        {wang2020particle}
\bibfield{author}{\bibinfo{person}{Bin Wang}, \bibinfo{person}{Bing Xue}, {and}
  \bibinfo{person}{Mengjie Zhang}.} \bibinfo{year}{2020}\natexlab{a}.
\newblock \showarticletitle{Particle Swarm Optimization for Evolving Deep
  Convolutional Neural Networks for Image Classification: Single-and
  Multi-Objective Approaches}.
\newblock In \bibinfo{booktitle}{{\em Deep Neural Evolution}}.
  \bibinfo{publisher}{Springer}, \bibinfo{pages}{155--184}.
\newblock


\bibitem[\protect\citeauthoryear{Wang, Xue, and Zhang}{Wang
  et~al\mbox{.}}{2020b}]%
        {wang2020surrogate}
\bibfield{author}{\bibinfo{person}{Bin Wang}, \bibinfo{person}{Bing Xue}, {and}
  \bibinfo{person}{Mengjie Zhang}.} \bibinfo{year}{2020}\natexlab{b}.
\newblock \showarticletitle{Surrogate-assisted Particle Swarm Optimisation for
  Evolving Variable-length Transferable Blocks for Image Classification}.
\newblock \bibinfo{journal}{{\em arXiv preprint arXiv:2007.01556\/}}
  (\bibinfo{year}{2020}).
\newblock


\bibitem[\protect\citeauthoryear{Xiao, Rasul, and Vollgraf}{Xiao
  et~al\mbox{.}}{2017}]%
        {xiao2017fashion}
\bibfield{author}{\bibinfo{person}{Han Xiao}, \bibinfo{person}{Kashif Rasul},
  {and} \bibinfo{person}{Roland Vollgraf}.} \bibinfo{year}{2017}\natexlab{}.
\newblock \showarticletitle{Fashion-mnist: a novel image dataset for
  benchmarking machine learning algorithms}.
\newblock \bibinfo{journal}{{\em arXiv preprint arXiv:1708.07747\/}}
  (\bibinfo{year}{2017}).
\newblock


\bibitem[\protect\citeauthoryear{Xie and Yuille}{Xie and Yuille}{2017}]%
        {xie2017genetic}
\bibfield{author}{\bibinfo{person}{Lingxi Xie} {and} \bibinfo{person}{Alan
  Yuille}.} \bibinfo{year}{2017}\natexlab{}.
\newblock \showarticletitle{Genetic cnn}. In \bibinfo{booktitle}{{\em
  Proceedings of the IEEE International Conference on Computer Vision}}.
  \bibinfo{pages}{1379--1388}.
\newblock


\bibitem[\protect\citeauthoryear{Zagoruyko and Komodakis}{Zagoruyko and
  Komodakis}{2016}]%
        {zagoruyko2016wide}
\bibfield{author}{\bibinfo{person}{Sergey Zagoruyko} {and}
  \bibinfo{person}{Nikos Komodakis}.} \bibinfo{year}{2016}\natexlab{}.
\newblock \showarticletitle{Wide Residual Networks}. In
  \bibinfo{booktitle}{{\em Proceedings of the British Machine Vision Conference
  (BMVC)}}. \bibinfo{pages}{87.1--87.12}.
\newblock


\bibitem[\protect\citeauthoryear{Zhong, Yan, Wu, Shao, and Liu}{Zhong
  et~al\mbox{.}}{2018}]%
        {zhong2018practical}
\bibfield{author}{\bibinfo{person}{Zhao Zhong}, \bibinfo{person}{Junjie Yan},
  \bibinfo{person}{Wei Wu}, \bibinfo{person}{Jing Shao}, {and}
  \bibinfo{person}{Cheng-Lin Liu}.} \bibinfo{year}{2018}\natexlab{}.
\newblock \showarticletitle{Practical block-wise neural network architecture
  generation}. In \bibinfo{booktitle}{{\em Proceedings of the IEEE Conference
  on Computer Vision and Pattern Recognition}}. \bibinfo{pages}{2423--2432}.
\newblock


\bibitem[\protect\citeauthoryear{Zoph and Le}{Zoph and Le}{2017}]%
        {zoph2016neural}
\bibfield{author}{\bibinfo{person}{Barret Zoph} {and} \bibinfo{person}{Quoc~V.
  Le}.} \bibinfo{year}{2017}\natexlab{}.
\newblock \showarticletitle{Neural Architecture Search with Reinforcement
  Learning}. In \bibinfo{booktitle}{{\em 5th International Conference on
  Learning Representations (ICLR)}}.
\newblock
\showURL{%
\url{https://arxiv.org/abs/1611.01578}}


\bibitem[\protect\citeauthoryear{Zoph, Vasudevan, Shlens, and Le}{Zoph
  et~al\mbox{.}}{2018}]%
        {zoph2018learning}
\bibfield{author}{\bibinfo{person}{Barret Zoph}, \bibinfo{person}{Vijay
  Vasudevan}, \bibinfo{person}{Jonathon Shlens}, {and} \bibinfo{person}{Quoc~V
  Le}.} \bibinfo{year}{2018}\natexlab{}.
\newblock \showarticletitle{Learning transferable architectures for scalable
  image recognition}. In \bibinfo{booktitle}{{\em Proceedings of the IEEE
  conference on computer vision and pattern recognition}}.
  \bibinfo{pages}{8697--8710}.
\newblock


\end{thebibliography}

\end{document}